\documentclass[runningheads]{llncs}

\usepackage{eccv}

\usepackage{eccvabbrv}

\usepackage{graphicx}
\usepackage{booktabs}

\usepackage[accsupp]{axessibility}  
\usepackage{amsfonts}       

\usepackage{subfiles} %
\usepackage{makecell} %
\usepackage{multirow} %
\usepackage{color,xcolor}
\usepackage{graphicx}
\usepackage{booktabs}
\usepackage{makecell}
\usepackage{colortbl}
\usepackage{pifont}%
\usepackage{algorithm} %
\usepackage{algorithmic} %
\usepackage{stix}   
\usepackage{ulem}
\usepackage[misc]{ifsym}

\usepackage{amsmath,amsfonts}       %
\usepackage{bm}
\usepackage{bbm}
\usepackage{multirow}
\usepackage{diagbox}
\usepackage{comment}
\usepackage{etoolbox}
\usepackage{wrapfig}
\usepackage[font=small]{caption}
\usepackage{tablefootnote}
\newtheorem{MyTheorem1}{Convolution Theorem}
\newtheorem{MyTheorem2}{Rayleigh’s Theorem}

\definecolor{shapecolor}{rgb}{0.0,0.5,0.0}

\definecolor{arylideyellow}{rgb}{0.91, 0.84, 0.42}

\usepackage[pagebackref,breaklinks,colorlinks,citecolor=eccvblue]{hyperref}

\usepackage{orcidlink}

\newcommand{\sysabbrev}{\textsc{M2}\xspace}

\newcommand{\dt}{\Delta}

\begin{document}

\title{SiMBA: Simplified Mamba-based Architecture for Vision and Multivariate Time series} 

\titlerunning{SiMBA: Simplified Mamba}

\author{Badri N. Patro\inst{1} \and
Vijay S, Agneeswaran\inst{1}}

\authorrunning{B. N. Patro et al.}

\institute{Microsoft 
\\
\email{\{badripatro,vagneeswaran\}@microsoft.com}}

\maketitle

\begin{abstract}
Transformers have widely adopted attention networks for sequence mixing and MLPs for channel mixing, playing a pivotal role in achieving breakthroughs across domains. However, recent literature highlights issues with attention networks, including low inductive bias and quadratic complexity concerning input sequence length.  State Space Models (SSMs) like S4 and others (Hippo, Global Convolutions, liquid S4, LRU, Mega, and Mamba), have emerged to address the above issues to help handle longer sequence lengths. 
Mamba, while being the state-of-the-art SSM, has a stability issue when scaled to large networks for computer vision datasets. We propose SiMBA, a new architecture that introduces Einstein FFT (EinFFT) for channel modeling by specific eigenvalue computations and uses the Mamba block for sequence modeling. Extensive performance studies across image and time-series benchmarks demonstrate that SiMBA outperforms existing SSMs, bridging the performance gap with state-of-the-art transformers. Notably, SiMBA establishes itself as the new state-of-the-art SSM on ImageNet and transfer learning benchmarks such as Stanford Car and Flower as well as task learning benchmarks as well as seven time series benchmark datasets. The project page is available on this website
~\url{https://github.com/badripatro/Simba}
 
\end{abstract}

\section{Introduction}
\label{sec:intro}

The evolution of language models in the technological landscape is transitioning from Large Language Models (LLMs) to the paradigm of Small Language Models (SLMs) inspired by cutting-edge SLM architectures like Mistral \cite{jiang2023mistral}, Phi \cite{li2023textbooks}, Orca \cite{mukherjee2023orca}, among others. At the heart of both LLMs and SLMs lies the power of transformers, where the layers are not only scaled but also exhibit scaling in both the token and channel modeling. This has made transformers the building blocks of LLMs and SLMs.

Transformers operate through two fundamental mixing directions: sequential modeling, involving the interaction of one token with another within the input sequence, and channel modeling, facilitating interactions within the channel dimension or across input features. Traditionally, multi-headed self-attention (MHSA) was employed by transformers for sequence modeling, but its computational complexity of $O(N^2)$ posed inefficiencies and performance challenges, particularly for longer sequences. To address this limitation, and recognize the need for handling extended input sequences in domains like genomics or protein folding, a novel approach emerged with the introduction of the Structured State Space model (S4) \cite{gu2021efficiently}. This model leverages state-space-based sequential modeling, offering enhanced efficiency and performance for processing longer input sequences. 

\begin{figure*}
\centering
  \includegraphics[width=0.85\linewidth]{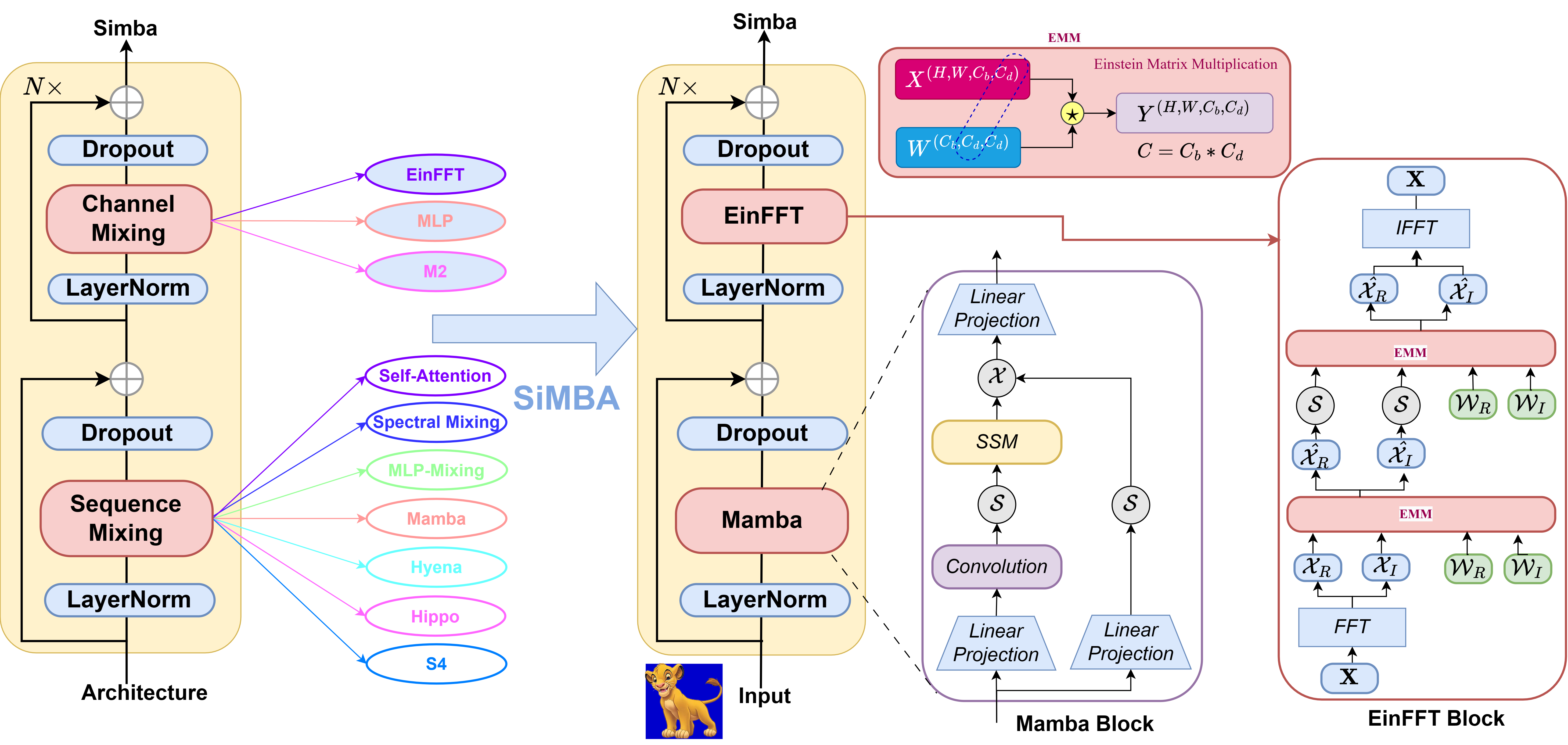}
  \caption{Simplified Mamba Based Architecture.}
  \label{fig:mamba_teaser}
  \vspace{-0.3in}
\end{figure*}

 However, S4 has been less effective in catering to modeling information-dense data, particularly in domains such as computer vision, and faces challenges in discrete scenarios like genomic data. In response to these limitations, several other SSMs, including Hippo\cite{albert2020hippo}, H3~\cite{fu2022hungry}, Hyena ~\cite{poli2023hyena}, RetNet\cite{sun2023retentive}, RWKV ~\cite{peng2023rwkv} and S4nd ~\cite{nguyen2022s4nd} and Gated State Spaces \cite{mehta2022long}, have been proposed in the literature to address the shortcomings of S4, specifically focusing on enhancing its ability to handle long-range dependencies. Additionally, global convolutions-based~\cite{fu2023flashfftconv} state-space models have been introduced as an alternative solution to mitigate issues related to long-range dependencies. Further expansions in the S4 model family include Liquid S4 models~\cite{hasani2021liquid} and a variant known as Mega~\cite{ma2022mega}, which interprets S4 through an exponential moving average approach. These advancements collectively contribute to a more versatile and effective suite of models for diverse applications.
Mamba \cite{gu2023mamba} is a selective state space sequence modeling technique proposed recently to address difficulties typical state space models had in handling long sequences efficiently. The typical state space models have trouble propagating or forgetting information in long sequences. Mamba handles this difficulty, by incorporating the current token in the state space, achieving in-context learning. Mamba in general addresses the two concerns of transformers, namely its lack of inductive bias by using CNN and its quadratic complexity by using input-specific state space models.

The current instantiation of Mamba has stability issues i.e. the training loss is not converging while scaling to large-sized networks on the ImageNet dataset. It is not clear as to why Mamba has this instability while scaling to large networks. This leads to the problem of vanishing/exploding gradients commonly observed in Mamba in general. Existing literature, such as Oppenheim and Verghese's work~\cite{oppenheim2010}, establishes that linear state space models exhibit stability when all eigenvalues of matrix A are negative real numbers. This motivates the need for a stable Mamba architecture, as presented in this paper, and specifically, we use Fourier Transforms followed by a learnable layer with non-linearity.  SiMBA also introduces residual connections with dropouts, which help in handling solving the instability issues in Mamba, as illustrated in Figure-\ref{fig:mamba_teaser}. This strategic application of learnable Fourier Transforms aims to manipulate the eigenvalues, ensuring they are negative real numbers. Thus, The proposed channel modeling technique, named EinFFT, is a distinctive contribution to the field. This is a unique contribution, as SSMs in the literature have not addressed channel modeling explicitly.

\begin{table}[htb]
    \caption{Overview of Large Vision Models for Image Recognition Task.}
    \scriptsize
    \label{tab:overview}
    \centering
    \begin{tabular}{lcccc} \toprule
    {Method} & Seqence Mixing & Channel Mixing & Models & \\ \midrule
    Convolution  &  ConvNet & MLP &  AlexNet, VGG, ResNet, RegNetY& \\
    Transformer   & Attention  & MLP & ViT, Deit, T2T, TnT & \\
  
    MLP-Mixer &  MLP  & MLP & Mlp-Mixer, GMLP, DynamicMLP & \\
    Spectral-Mixer &  FFT/Wavelet  & MLP &GFNet, AFNO,WaveMix & \\
     State Space &  SSM  & - & S4, Hyena, Hippo, H3, Mamba &\\  
  
     \midrule
    Conv-Transformer   & Attention+ ConvNet  & MLP & Swin, CvT, CMT, CSwin & \\
    Spectral-Transformer &  FFT +Attention  & MLP & SiMBA, SVT, WaveViT & \\

    \midrule
       SiMBA &  Mamba  & EinFFT & SiMBA &\\
    
    \bottomrule 
    \end{tabular}
\end{table}

Our architectural investigation explores diverse state space models, including S4\cite{gu2021efficiently}, Hippo\cite{albert2020hippo}, Hyena\cite{poli2023hyena}, and Mamba\cite{gu2023mamba}, alongside attention models such as DeIT\cite{touvron2021training}, ViT\cite{dosovitskiy2020image}, and Swin\cite{liu2022swin}, as well as spectral models like GFNet\cite{rao2021global}, and AFNO\cite{guibas2021efficient} for sequence modeling. Additionally, we explore various channel modeling alternatives, including MLP, and Monarch Mixer, and introduce a novel technique called EinFFT. Through extensive experimentation, we have identified the most efficient and streamlined state space architecture to date, named SiMBA. This novel architecture incorporates Mamba for sequence modeling and introduces EinFFT as a new channel modeling technique. SiMBA effectively addresses the instability issues observed in Mamba when scaling to large networks. The architectural alternatives explored for large-scale sequence modeling are depicted in Figure-\ref{fig:mamba_teaser}. Table-\ref{tab:overview} provides an overview of large vision models used for image recognition tasks, categorizing them based on their sequence mixing and channel mixing techniques. It highlights a diverse range of models, including those based on convolutional models, transformers models, MLP-mixers, spectral-mixers models, and state space methods. Additionally, it introduces hybrid models combining convolution with transformers or spectral approaches. Lastly, it presents SiMBA, a novel model utilizing Mamba for sequence mixing and EinFFT for channel mixing.

The contribution of this paper is as follows:
\begin{itemize}
\item 
EinFFT: A new technique for channel modeling known as EinFFT is proposed, which solves the stability issue in Mamba. This uses Fourier transforms with non-linearity to model eigenvalues as negative real numbers, which solves instability \cite{oppenheim2010}. We validate this technique on two data modalities time series and ImageNet dataset. 
\item 
SiMBA: We propose an optimized Mamba architecture for computer vision tasks, known as SiMBA which uses EinFFT for channel modeling and Mamba for token mixing to handle inductive bias and computational complexity. We also show the importance of different architectural elements of SiMBA with ablation studies including residual connections and dropouts.
\item 
Performance Gap: It must be noted that SiMBA is the first SSM to close the performance gap with state-of-the-art attention-based transformers on the ImageNet dataset and six standard time series datasets. We show with extensive performance analysis how SiMBA achieves state-of-art performance compared to V-Mamba and Vision Mamba in the vision domain. We have shown the generalization capability of SiMBA architecture in other domains like time series to handle long sequences.
 \item 
We show the performance of SiMBA on transfer learning datasets like CIFAR, Stanford Car, and Flower. We also show that SiMBA achieves comparable performance even in other tasks such as instance segmentation with the MS COCO dataset.
\end{itemize}


\section{Related work}
At the heart of SiMBA's transformer-based architecture lies a combination of multi-headed self-attention and a multi-layer perceptron (MLP). When comparing SiMBA with attention-based transformers like ViT\cite{dosovitskiy2020image}, DeIT\cite{touvron2021training}, Swin\cite{liu2021swin}, CSwin\cite{dong2022cswin}, and CVT\cite{wu2021cvt}, it becomes evident that ViT, although pioneering attention-based transformers in computer vision, faced challenges due to its quadratic attention complexity ($O(N^2)$). Subsequent models like DeIT, Swin, TvT, CVT, and CSwin aimed at refining transformer performance in computer vision and specific NLP tasks. Alternatively, MLP Mixer architectures such as MLP Mixers\cite{tolstikhin2021mlp} and DynamicMLP\cite{wang2022dynamixer} sought to replace attention networks with MLP to address these challenges. Efforts to mitigate attention complexity led to the development of transformers like GFNet\cite{rao2021global}, AFNO\cite{guibas2021efficient}, FNet\cite{lee2021fnet}, and FourierFormer\cite{nguyen2022fourierformer}, incorporating Fourier transforms to reduce complexity to $O(N\log(N))$. Despite this reduction, these models demonstrated a performance gap compared to state-of-the-art attention-based transformers. The emergence of transformers like SVT\cite{patro2023scattering}, WaveViT\cite{yao2022wave}, and SpectFormer\cite{patro2023spectformer}, combining initial spectral layers with deeper attention mechanisms, surpassed the performance of both attention-based and MLP-based transformers. This comprehensive comparative analysis highlights SiMBA's distinctive position in the evolving landscape of transformer architectures.

Attention-based transformers encounter limitations in modeling long input sequences, especially when dependencies extend beyond the attention window size. This constraint proves crucial in applications such as high-resolution imagery analysis and genomics. S4 pioneered state space models to address this issue by reducing complexity to $O(N\log(N))$, enabling the handling of long-range dependencies in tasks like the long-range arena path-X\cite{lrabench}. Subsequent efforts, including Hippo and Long Convolutions\cite{fu2023simple}, aimed to enhance state space models' efficiency but demonstrated a performance gap compared to state-of-the-art transformers. Hyena\cite{poli2023hyena}, H3\cite{fu2022hungry}, and related models further improved state space models' effectiveness but faced challenges due to a lack of consideration for the current input. Mamba attempted to bridge this gap by parameterizing the current input in the state space model. However, state-space models often encounter instability issues due to their sensitivity to recurrent dynamics.

Vision Mamba\cite{zhu2024vision} and V-Mamba\cite{liu2024vmamba} adapted the Mamba architecture for computer vision tasks, utilizing bi-directional and visual state space models. However, the performance study section reveals a performance gap between Vision Mamba, V-Mamba, and state-of-the-art transformer models like SpectFormer\cite{patro2023spectformer}, SVT\cite{patro2023scattering}, WaveViT\cite{yao2022wave}, and Volo\cite{yuan2022volo}. SiMBA addresses this gap by incorporating Mamba for token mixing, replacing attention networks, and leveraging Einstein FFT (EinFFT) for channel mixing. SiMBA introduces the Einstein blending method for channel mixing, offering a novel approach without the constraints of requiring perfect square dimensions for sequence length N and channel dimensions. Furthermore, SiMBA adopts the pyramid version of the transformer architecture, providing a significant performance boost compared to vanilla state space models. While many state space models reduce complexity to $O(N\log(N))$, they often fall short of achieving the performance levels seen in state-of-the-art attention-based transformers.

\vspace{-0.1in}
\section{Method }
\vspace{-0.1in}
In this study, we introduce EinFFT, a novel approach for frequency-domain channel mixing utilizing Einstein Matrix multiplication. EinFFT is specifically designed for complex number representations of frequency components, enabling the effective capture of key patterns in Image patch data with a global view and energy compaction. It must be noted that EinFFT is also applicable for other sequence data modalities like time series or speech or even text data. We have validated EinFFT-based SiMBA for image and time series benchmark datasets.

\subsection{Channel Mixing: Einstein FFT (EinFFT)} \label{sec:fremlp}
The Channel Mixing component of SiMBA (EinFFT) encompasses three main components: Spectral Transformation, Spectral Gating Network using Einstein Matrix multiplication, and Inverse Spectral Transformation as depicted in Figure \ref{fig:mamba_teaser}.
\subsubsection{Spectral Transformation}
For a given input function  $\mathbf{f}(x)$ and its corresponding frequency domain conversion function $\mathcal{F}(k)$, by using  Discrete Fourier Transform ($\operatorname{DFT}$). Let \(\mathcal{F}\) denote the Fourier transform of a function $f(x)$ and $\mathcal{F}^{-1}$ its inverse then
    \begin{equation}\label{fft}
    \mathcal{F}(k) = \int_{D} f(x) e^{- j2 \pi k x } \mathrm{d}x = \int_{D} {f}(x) \cos(2\pi kx)\mathrm{d}x + j \int_{D} {f}(x) \sin(2\pi kx)\mathrm{d}x 
    \end{equation}
Here, $k$ is the frequency variable, $x$ is the spatial variable, and $j$ is the imaginary unit. The real part of $\mathcal{F}$ is denoted as $Re(\mathcal{F})$, and the imaginary part as $Im(\mathcal{F})$. The complete conversion is expressed as $\mathcal{F}=Re(\mathcal{F})+jIm(\mathcal{F})$. Fourier transform is employed to decompose the input signal into its constituent frequencies. This allows the identification of periodic or aperiodic patterns which are crucial for image recognition or detection tasks.

\begin{MyTheorem1}
The convolution theorem states that the Fourier transform of the convolution of two functions in the spatial domain is equal to the pointwise product of their Fourier transforms. Mathematically, if \( f(x) \) and \( g(x) \) are two functions in the spatial domain, and \( F(u) \) and \( G(u) \) are their respective Fourier transforms, then the convolution theorem can be expressed as:
\[ \mathcal{F}\{f * g\}(u) = F(u) \cdot G(u) \]
where:\( \mathcal{F} \) denotes the Fourier transform operator. \(* \) denotes the convolution operator.

\end{MyTheorem1}

\begin{MyTheorem2}

Rayleigh's Theorem states that the total energy (or power) of a signal in the spatial domain is equal to the total energy in its frequency domain representation.  Rayleigh's Theorem for continuous signals, and is used to express the equivalence of energy between an image patch in the spatial domain (\(|\mathbf{g}(x)|^2\)) and its representation in the frequency domain (\(|\mathcal{G}(f)|^2\)) as defined as:
\[ \int_{-\infty }^{\infty} |\mathbf{g}(x)|^2 \, dx = \int_{-\infty }^{\infty} |\mathcal{G}(f)|^2 \, df \]
Here, $\mathcal{G}(f)=\int_{-\infty }^{\infty}\mathbf{g}(x) e^{-j2\pi fx} \mathrm{d}x$, where $x$ represents the channel dimension, and $f$ denotes the frequency dimension.
 Rayleigh's Theorem describes the conservation of energy between the spatial domain and the frequency domain. It indicates that the integral of the squared magnitude of the signal in the time domain is equal to the integral of the squared magnitude of its Fourier transform in the frequency domain.

The implications of this theorem, emphasize that if the majority of the energy of an input image is concentrated in a limited number of frequency components, then an accurate representation of the image can be achieved by focusing on those specific components. This is a key insight from the theorem, emphasizing the concentration of energy in certain frequency components and the potential benefits of selectively considering those components for more efficient and informative representations in the frequency domain.

\end{MyTheorem2}

\subsubsection{Frequency-domain Channel Mixing}
For a given input $\mathbf{X}  \in \mathbb{R}^{N \times D}$ and its corresponding frequency domain conversion $\mathcal{X} \in \mathbb{R}^{N \times D}$, by using Fourier transform. The convolution theorem states that  the Fourier transform of the convolutions on $\mathbf{X}$ with its kernel in the spatial domain can be equivalently represented as the 
product operations of their frequency-domain representations.  This equivalence is expressed as follows:
\begin{equation}
     \mathcal{F}(\mathbf{X}\ast W+B) =\mathcal{F}(\mathbf{X}) \cdot \mathcal{F}(W)+\mathcal{F}(B) =\mathcal{X}\mathcal{W}+\mathcal{B}
\end{equation}
where $\ast$ denotes circular convolution, $\mathcal{W}$ and $\mathcal{B}$ represent the complex number weight and bias in the frequency domain, while $W$ and $B$ denote the weight and bias in the spatial domain, and $\mathcal{F}$ signifies the Discrete Fourier Transform ($\operatorname{DFT}$). The output DFT operation is a complex number value $\mathcal{X} \in \mathbb{R}^{N \times D}$, multiplied with a complex number weight matrix $\mathcal{W} \in \mathbb{R}^{D \times D}$ and added with a complex number bias $\mathcal{B}\in \mathbb{R}^{D}$ using the Spectral Gating Network (SGN).  To reduce complexity we perform Einstein Matrix Multiplication (EMM) to have an efficient block diagonal matrix. The SGN is expressed by the following formulation:
\begin{equation} \label{equ:sgn}
    {h}^{\ell} = \sigma({h}^{\ell-1}\mathcal{W}^{\ell}+\mathcal{B}^{\ell}), {h}^{0} = \mathcal{X}    
\end{equation}

Here, $h^{\ell} \in \mathbb{R}^{N \times D}$ represents the final output, $\ell$ denotes the $\ell$-th layer, and $\sigma$ is the activation function. Considering both $\mathcal{X}$ and $\mathcal{W}$ as complex numbers, we extend Equation (\ref{equ:sgn}) by employing the multiplication rule for complex numbers. The extended formulation is as follows:
\begin{equation}\label{equ:complex_multiply}
\begin{split}
    Re(h)^{\ell} &=\textit{EMM}(Re(h^{\ell-1})\mathcal{W}^{\ell}_r) - \textit{EMM}(Im(h^{\ell-1})\mathcal{W}^{\ell}_i )+\mathcal{B}^{\ell}_r \\    
   Im(h)^{\ell} &= \textit{EMM}(Re(h^{\ell-1})\mathcal{W}^{\ell}_i)+ \textit{EMM}(Im(h^{\ell-1})\mathcal{W}^{\ell}_r)+\mathcal{B}^{\ell}_i\\
    h^{\ell}&=  \sigma(Re(h)^{\ell} )+ j  \sigma(Im(h)^{\ell})
    \end{split}
\end{equation}

Here, $\mathcal{W}^{\ell}=\mathcal{W}^{\ell}_r+j\mathcal{W}^{\ell}_i$ and $\mathcal{B}^{\ell}=\mathcal{B}^{\ell}_r+j\mathcal{B}^{\ell}_i$ represent the real and imaginary parts, respectively. The implications of this theorem, emphasize that if the energy of an image patch is concentrated in a limited number of frequency components, then 
with the help of Rayleigh's Theorem a representation focusing on those specific frequency components can accurately capture the signal. this can be modeled with the help of a learnable spectral gating mechanism. We also apply a non-linear activation function to handle stability in the sequence modeling block (Mamba) by allowing a specific Eigenvalue that is required for convergence. We then apply EMM between input $\mathbf{I}$ and the weight  $W\in \mathbb{R}^ {C_b \times C_d \times C_d}$ along the channel dimensions, where the channels are arranged in number of blocks and number of subchannels in each block to make it block diagonal matrix ( $\mathbf{I}$ from $\mathbb{R}^{ N \times C}$ to $\mathbb{R}^{ N \times C_b\times C_d}$, where $C= C_b \times C_d $, and $b << d$). This results in a blended feature tensor $Y \in \mathbb{R}^{ N \times C_b \times C_d}$ as defined where $N = H\times W$. The formula for EMM is:
$$\mathbf{Y}^{ N \times C_b \times C_d}= \mathbf{I}^{ N \times C_b \times C_d} \boxast  \mathbf{W}^ {C_b \times C_d \times C_d} $$ Where $\boxast$ represents an Einstein matrix multiplication

The channel mixing in the frequency domain, denoted as $\operatorname{EinFFT}$, is performed by separately computing the real and imaginary parts of frequency components.  Subsequently, these parts are stacked to form a complex number, with the final result obtained by concatenating the real and imaginary parts. This methodology enhances the extraction of patterns while ensuring efficient utilization of computational resources.

\subsubsection{Inverse Spectral Transformation}: After learning in the frequency domain, $\mathcal{F}$ can be converted back into the time domain using the inverse conversion formulation:

    \begin{equation}\label{ifft}
        f(x) = \int_{D} \mathcal{F}(k) e^{j2\pi kx}\mathrm{d}f = \int_{D} (Re(\mathcal{F}(k))+jIm(\mathcal{F}(k)) e^{j2\pi kx}\mathrm{d}f
    \end{equation}
 
The frequency spectrum, expressed as a combination of $\cos$ and $\sin$ waves, aids in discerning prominent frequencies and periodic patterns in time series signals. The terms $\operatorname{FFT}$ and $\operatorname{IFFT}$ are used for brevity. The Spectral Transformation stage leverages the Fourier transform for enhanced signal analysis, capturing important periodic patterns crucial for the Image classification task.

\subsubsection{EinFFT Architecture for SiMBA}
Our method, EinFFT, leverages frequency-domain channel mixing through the application of Einstein Matrix multiplication on complex number representations. This enables the extraction of intricate data patterns with enhanced global visibility and energy concentration.

Considering channel dependencies is crucial for accurate class prediction as it allows the model to capture interactions and correlations between different variables. The frequency channel learner facilitates communication between different channels, and learning channel dependencies by sharing the same weights across $L$ timestamps. For the $l$-th timestamp $\mathbf{X}_c^{:,(l)} \in \mathbb{R}^{N \times d}$, the frequency channel learner operates as follows:

\begin{equation}
\begin{split}
     \mathcal{X}^{:,(l)}=\mathcal{X}_{R}^{:,(l)} + j \mathcal{X}_{I}^{:,(l)}&= \operatorname{FFT}(\mathbf{X}_c^{:,(l)})  \\
     Re(h)^{\ell}+j Im(h)^{\ell} &= \operatorname{EMM}_{}(\mathcal{X}^{:,(l)}, \mathcal{W}^{c}, \mathcal{B}^{c}) \text{ as in eq-\ref{equ:complex_multiply}}\\
     \mathcal{Y}^{:,(l)}=\mathcal{Y}_{R}^{:,(l)} +j  \mathcal{Y}_{I}^{:,(l)} &= \sigma(Re(h)^{\ell}) + j \sigma(Im(h)^{\ell})\\
        \mathcal{Z}^{:,(l)}=\mathcal{Z}_{R}^{:,(l)} +j  \mathcal{Z}_{I}^{:,(l)} &= \operatorname{EMM}_{}(\mathcal{Y}^{:,(l)}, \mathcal{W}^{c}, \mathcal{B}^{c}) \text{ as in eq-\ref{equ:complex_multiply}}\\
     \mathbf{Z}_{c}^{:,(l)} &= \operatorname{IFFT}_{}(\mathcal{Z}_{R}^{:,(l)} +j  \mathcal{Z}_{I}^{:,(l)}) 
\end{split}
\end{equation}

Here, $\mathcal{X}^{:,(l)} \in \mathbb{R}^{\frac{N}{2} \times d}$ represents the frequency components of $\mathbf{X}_c^{:,(l)}$. The operations $\operatorname{FFT}_{}$ and $\operatorname{IFFT}_{}$ are performed along the channel dimension. $\operatorname{EinFFT}$ refers to the frequency-domain channel Mixing introduced, taking $\mathcal{W}^{c}$ and $\mathcal{B}^{c}$ as the complex number weight matrix and biases, respectively. The output $\mathcal{Z}_{}^{:,(l)} \in \mathbb{R}^{\frac{N}{2} \times d}$ of the EinFFT is then converted back to the time domain as $\mathbf{Z}^{:,(l)} \in \mathbb{R}^{N \times d}$. Finally, the outputs $\mathbf{Z}^{:,(l)}$ across $L$ timestamps are ensembled to produce the overall output $\mathbf{Z}_t \in \mathbb{R}^{N \times L \times d}$.

\subsection{Sequence Modeling: Mamba based SSM}\label{sec:ss-continuous}

To model a large sequence we use state space models instead of Multi-headed self-attention due to its complexity. The state space model\cite{gu2021efficiently,gu2023mamba} is commonly known as a linear time-invariant system that map the input stimulation $x(t) \in \mathcal{R}^L $ to a response $y(t) \ $ through a hidden space $h(t) \in \mathcal{R}^N $. Structured state space sequence models (S4) are a recent class of sequence models for deep learning that are broadly related to RNNs, CNNs, and classical state space models. Mathematically,  The Continuous-time Latent State spaces can be modeled as linear ordinary differential equations that use evolution parameter $A \in \mathcal{R}^{N\times N} $ and projection parameter $B \in \mathcal{R}^{N\times 1} $ and $C \in \mathcal{R}^{N\times 1} $ as follows:

\begin{equation}
  \label{eq:1}
  \begin{aligned}
    x'(t) &= \bm{A}x(t) + \bm{B}u(t) \\
    y(t) &= \bm{C}x(t) + \bm{D}u(t)
  \end{aligned}
\end{equation}

\subsubsection{Discrete-time SSM: }
\label{sec:ss-recurrent}
.
The discrete form of SSM uses a time-scale parameter $\dt$ to transform continuous parameters A, B, and C to discrete parameters $\Bar{A}, \Bar{B}$ and $ \Bar{C} $ using fixed formula  $\Bar{A}=f_{A}(\dt,A), \Bar{B}=f_{B}(\dt,A,B)$. The pair $f_{A},f_{B}$ is the discretization rule that uses a zero-order hold (ZOH)  for this transformation. The equations are as follows.

\begin{equation}
  \label{eq:2}
\begin{aligned}
  x_{k} &= \bm{\overline{A}} x_{k-1} + \bm{\overline{B}} u_k &
  \bm{\overline{A}} &= (\bm{I} - \dt/2 \cdot \bm{A})^{-1}(\bm{I} + \dt/2 \cdot \bm{A}) &
  \\
  y_k &= \bm{\overline{C}} x_k &
  \bm{\overline{B}} &= (\bm{I} - \dt/2 \cdot \bm{A})^{-1} \dt \bm{B} &
  \bm{\overline{C}} &= \bm{C}
  .
\end{aligned}
\end{equation}

\subsubsection{ Convolutional Kernel Representation}
\label{sec:ss-convolution}
The discretized form of recurrent SSM in equation-\ref{eq:2} is not practically trainable due to its sequential nature. To get efficient representation, we model continuous convolution as discrete convolution as it is a linear time-invariant system. 
For simplicity let the initial state be \( x_{-1} = 0 \).
Then unrolling \eqref{eq:2} explicitly yields: %
\begin{align*}
  x_0 &= \bm{\overline{B}} u_0 &
  x_1 &= \bm{\overline{A}} \bm{\overline{B}} u_0 + \bm{\overline{B}} u_1 &
  x_2 &= \bm{\overline{A}}^2 \bm{\overline{B}} u_0 + \bm{\overline{A}} \bm{\overline{B}} u_1 + \bm{\overline{B}} u_2 & \dots
  \\
  y_0 &= \bm{\overline{C}} \bm{\overline{B}} u_0 &
  y_1 &= \bm{\overline{C}} \bm{\overline{A}} \bm{\overline{B}} u_0 + \bm{\overline{C}} \bm{\overline{B}} u_1 &
  y_2 &= \bm{\overline{C}} \bm{\overline{A}}^2 \bm{\overline{B}} u_0 + \bm{\overline{C}} \bm{\overline{A}} \bm{\overline{B}} u_1 + \bm{\overline{C}} \bm{\overline{B}} u_2
  & \dots
\end{align*}
This can be vectorized into a convolution \eqref{eq:convolution} with an explicit formula for the convolution kernel \eqref{eq:krylov}.
\begin{equation}
  \label{eq:convolution}
  \begin{split}
    y_k &= \bm{\overline{C}} \bm{\overline{A}}^k \bm{\overline{B}} u_0 + \bm{\overline{C}} \bm{\overline{A}}^{k-1} \bm{\overline{B}} u_1 + \dots + \bm{\overline{C}} \bm{\overline{A}} \bm{\overline{B}} u_{k-1} + \bm{\overline{C}}\bm{\overline{B}} u_k
    \\
    y &= \bm{\overline{K}} \ast u %
    .
  \end{split}
\end{equation}
\begin{equation}%
  \label{eq:krylov}
  \bm{\overline{K}} \in \mathbb{R}^L :=
  \mathcal{K}_L(\bm{\overline{A}}, \bm{\overline{B}}, \bm{\overline{C}}) := \left(\bm{\overline{C}} \bm{\overline{A}}^i \bm{\overline{B}}\right)_{i \in [L]} = (\bm{\overline{C}}\bm{\overline{B}}, \bm{\overline{C}}\bm{\overline{A}}\bm{\overline{B}}, \dots, \bm{\overline{C}}\bm{\overline{A}}^{L-1}\bm{\overline{B}})
  .
\end{equation}
\normalsize
 \( \bm{\overline{K}} \) in equation \eqref{eq:convolution} can be represented as a single (non-circular) convolution which can be computed very efficiently with FFTs.However, computing \( \bm{\overline{K}} \) in \eqref{eq:krylov} is non-trivial and is modelled as a \( \bm{\overline{K}} \) the \textbf{SSM convolution kernel} or filter.

Specifically, we model input sequence using the state-of-the-art state space model Mamba\cite{gu2023mamba}. Mamba identifies a critical weakness in existing models: their inability to perform content-based reasoning.
To address this, Mamba introduces selective state spaces (SSMs) that allow the model to selectively propagate or forget information along the sequence length dimension based on the current token.  While we apply the Mamba block for the vision task, we face the problem of stability issues (loss convergence issue) compared to other models like S4 or Hippo.  We are providing one type of solution for the instability issue by preserving only negative eigenvalue. To perform this we need an extra module which we call the channel mixing, which was missing in the mamba block. We combine the channel mixing module with the Sequence mixing module and make a Simplified Mamba Based Architecture (SiMBA) as shown in the figure-\ref{fig:mamba_teaser}.  The input token sequence $\mathbf{X}_{\mathtt{l}-1}$ is first normalized by the normalization layer. Next, we linearly project the normalized sequence to the $\mathbf{x}$ and $\mathbf{z}$ with dimension size $E$. Then, we process the $\mathbf{x}$ from the forward and backward directions. 
For each direction, we first apply the 1-D convolution to the $\mathbf{x}$ and get the $\mathbf{x}'_{o}$. We then linearly project the $\mathbf{x}'_{o}$ to the $\mathbf{B}_{o}$, $\mathbf{C}_{o}$, $\mathbf{\Delta}_{o}$, respectively. 
The $\mathbf{\Delta}_{o}$ is then used to transform the $\overline{\mathbf{A}}_{o}$, $\overline{\mathbf{B}}_{o}$, respectively.

 $\mathbf{SiMBA}$  uses Mamba block for sequence modeling, $\mathtt{L}$ denotes the number of layers, and $\mathbf{Norm}$ stands for the normalization layer and EinFFT for channel modeling. The process involves applying the Mamba block to the previous time step $\mathbf{X}_{\mathtt{l}-1}$, followed by a residual connection and dropout ($\mathbf{DP}$). The resulting tensor is then normalized ($\mathbf{Norm}$) to obtain the final sequence vector. The subsequent operation involves applying EinFFT, which is the proposed frequency-domain channel mixing operation. Finally, the tensor undergoes another dropout and is added to the previous state $\mathbf{X}_l$.

SiMBA illustrates an important trade-off between performance and scalability. Mamba by itself may have stability issues for large networks. Mamba combined with MLP for channel mixing bridges the performance gap for small-scale networks, but may have the same stability issues for large networks. Mamba combined with EinFFT solves stability issues for both small-scale and large networks. There may still be a performance gap with state-of-the-art transformers for large networks, which we shall address in future work.

\begin{table}[!htb]
\centering
\scriptsize
\begin{tabular}{c|ccc|c}
\toprule
Method & Image Size & \#Param. & FLOPs  & Top-1 acc. \\

\toprule
\multicolumn{4}{c}{\textbf{Convnets}} \\
\midrule

ResNet-101 & $224^{2}$& 45M&-  & 77.4\\
\rowcolor{gray!15}RegNetY-8G~\cite{radosavovic2020designing} & 224$^2$ & 39M & 8.0G  & 81.7 \\
\midrule
ResNet-152 &$224^{2}$ & 60M&-  & 78.3\\
\rowcolor{gray!15}RegNetY-16G~\cite{radosavovic2020designing} & 224$^2$ & 84M & 16.0G  & 82.9 \\
\toprule
\multicolumn{4}{c}{\textbf{Transformers}} \\

\midrule
DeiT-S~\cite{touvron2021training} & 224$^2$ & 22M & 4.6G  & 79.8 \\
Swin-T~\cite{liu2022swin} & 224$^2$ & 29M & 4.5G  & 81.3 \\
EffNet-B4~\cite{li2022efficientformer} & 380$^2$ & 19M & 4.2G  & 82.9\\
WaveViT-H-S$^\star$~\cite{yao2022wave} & 224$^2$ & 22.7M & 4.1G  & 82.9\\
\rowcolor{gray!15}SVT-H-S$^\star$~\cite{patro2023scattering} & 224$^2$ & 22M & 3.9G  & 84.2\\
\midrule

EffNet-B5~\cite{li2022efficientformer} & 456$^2$ & 30M & 9.9G & 83.6 \\
Swin-S~\cite{liu2022swin} & 224$^2$ & 50M & 8.7G  & 83.0 \\
CMT-B ~\cite{guo2022cmt}& 224$^2$&45M&9.3G& 84.5\\

MaxViT-S~\cite{tu2022maxvit}& 224$^2$& 69M& 11.7G& 84.5\\
iFormer-B\cite{si2022inception}& 224$^2$ & 48M& 9.4G & 84.6 \\

Wave-ViT-B$^\star$  \cite{yao2022wave}& 224$^2$& 33M& 7.2G & 84.8  \\
\rowcolor{gray!15}{SVT-H-B$^\star$}\cite{patro2023scattering} & 224$^2$& {32.8M} & {6.3G}& {85.2} \\
\midrule
ViT-b + Monarch~\cite{fu2024monarch} &224$^2$ &33M &-& 78.9 \\
\sysabbrev-ViT-b~\cite{fu2024monarch}  &224$^2$ &45M  &- & {79.5}  \\
DeiT-B~\cite{touvron2021training} & 224$^2$ & 86M & 17.5G  & 81.8 \\
Swin-B~\cite{liu2022swin} & 224$^2$ & 88M & 15.4G  & 83.5 \\
\sysabbrev-Swin-B~\cite{fu2024monarch}  & 224$^2$&50M & -& 83.5  \\ 
EffNet-B6~\cite{li2022efficientformer} & 528$^2$ & 43M & 19.0G  & 84.0 \\
MaxViT-B ~\cite{tu2022maxvit}& 224$^2$& 120M& 23.4G&85.0\\
VOLO-D2$^\star$ \cite{yuan2022volo}& 224$^2$ & 58M& 14.1G & 85.2 \\
VOLO-D3$^\star$ \cite{yuan2022volo} & 224$^2$ & 86M& 20.6G & 85.4  \\
Wave-ViT-L$^\star$ \cite{yao2022wave} & 224$^2$ & 57M& 14.8G & 85.5  \\

\rowcolor{gray!15}{SVT-H-L$^\star$}\cite{patro2023scattering} & 224$^2$& {54.0M} & {12.7G} & {85.7} \\

\toprule
\multicolumn{4}{c}{\textbf{SSMs}} \\

\midrule
Vim-Ti\cite{zhu2024vision} &$224^{2}$ & 7M&- & 76.1 \\ %
VMamba-T\cite{liu2024vmamba} & 224$^2$ & 22M & 5.6G  & 82.2 \\
\rowcolor{gray!15}SiMBA-S(Monarch) (Ours) & 224$^2$ & 18.5M & 3.6G  & 81.1 \\
\rowcolor{gray!15}SiMBA-S(EinFFT) (Ours) & 224$^2$ & 15.3M &2.4G  & 81.7 \\
\rowcolor{gray!15}SiMBA-S(MLP) (Ours) & 224$^2$ & 26.5M & 5.0G & 84.0 \\
\midrule
Vim-S\cite{zhu2024vision} & $224^{2}$ & 26M &-& 80.5 \\
VMamba-S\cite{liu2024vmamba} & 224$^2$ & 44M & 11.2G  & 83.5 \\
\rowcolor{gray!15}SiMBA-B(Monarch) (Ours) & 224$^2$ & 26.9M & 5.5G  & 82.6 \\
\rowcolor{gray!15}SiMBA-B(EinFFT) (Ours) & 224$^2$ & 22.8M &4.2G  & 83.5 \\
\rowcolor{gray!15}SiMBA-B(MLP) (Ours) & 224$^2$ & 40.0M & 9.0G  & 84.7 \\
\midrule
HyenaViT-B~\cite{fu2024monarch}  &224$^2$ &88M &- & 78.5 \\
S4ND-ViT-B~\cite{nguyen2022s4nd} & 224$^2$ & 89M & -  & 80.4 \\
VMamba-B\cite{liu2024vmamba} & 224$^2$ & 75M & 18.0G & 83.2 \\

\rowcolor{gray!15}SiMBA-L(Monarch) (Ours) & 224$^2$ & 42M & 8.7G & 83.8 \\
\rowcolor{gray!15}SiMBA-L(EinFFT) (Ours) & 224$^2$ & 36.6M & 7.6G & 84.4 \\
\rowcolor{gray!15}SiMBA-L(MLP)$^\dagger$ (Ours) & 224$^2$ & 66.6M & 16.3G & 49.4 \\
\bottomrule
\end{tabular}
\normalsize
\caption{\textbf{SOTA on ImageNet-1K}The table shows the performance of various vision backbones on the ImageNet1K\cite{deng2009imagenet} dataset for image recognition tasks.  $\star$ indicates additionally trained with the Token Labeling~\cite{wang2022scaled} for patch encoding. We have grouped the vision models into three categories based on their GFLOPs (Small, Base, and Large). The GFLOP ranges: Small (GFLOPs$<$5), Base (5$\leq$GFLOPs$<$10), and Large (10$\leq$GFLOPs$<$30). $^\dagger$ indicates that instability in the training SSM.}
\label{tab:imagenet_sota}
 \vspace{-0.40in}
\end{table}

\section{Experiment }
 We conducted a comprehensive evaluation of SiMBA on key computer vision tasks, including image recognition, and instance segmentation as well as on other data modalities such as time series. Our assessments for SiMBA model on standard datasets involved the following: a) Training and evaluating ImageNet1K~\cite{deng2009imagenet} from scratch for the image recognition task. b) Ablation analysis comparing SiMBA with various state space architectures for sequence modeling and various architectures for channel modeling. c) Evaluation of SiMBA on other data modalities such as time series datasets - namely the multi-variate time series benchmark \cite{autoformer}. d) Transfer learning on CIFAR-10~\cite{krizhevsky2009learning}, CIFAR-100~\cite{krizhevsky2009learning}, Stanford Cars~\cite{krause20133d}, and Flowers-102~\cite{nilsback2008automated} for image recognition.
e) Fine-tuning SiMBA for downstream instance segmentation tasks. All experiments were conducted on a hierarchical transformer architecture, currently a state-of-the-art model, with an image size of $224 \times 224 \times 3$.  

\subsection{SOTA Comparison on ImageNet 1K}
We conducted performance evaluations on the ImageNet 1K dataset, comprising 1.2 million training images and 50,000 validation images distributed across 1000 categories. Our results, presented in Table \ref{tab:imagenet_sota}, categorize architectures based on their size: small (<5 GFlops), base (5-10 GFlops), and large (>10 GFlops). In the small category, SiMBA(MLP) demonstrates remarkable performance with an 84.0\% top-1 accuracy, surpassing prominent convolutional networks like ResNet-101 and ResNet-152, as well as leading transformers such as EffNet, ViT, Swin, and DeIT. Our comparison extends to state space models (SSMs) like S4ND, HyenaViT\cite{poli2023hyena}, VMambaa\cite{liu2024vmamba}, and Vim\cite{zhu2024vision}, where SiMBA not only outperforms them significantly but also maintains comparable complexity. Specifically, SiMBA-S(EinFFT) and SiMBA-S(MLP) stand out among other small models, including SSMs, small convolutional networks, and transformers, achieving accuracies of 82.7\% and 84.0\%, respectively. Furthermore, it surpasses competitor transformers like Wave-ViT-S, Max-ViT-T, and iFormer-S with lower GFlops and parameters while being on par with SpectFormer. Notably, SiMBA-S outperforms SSMs like Vim-T (76.1\%) and V-Mamba-T (82.2\%). We compare among our variants such as SiMBA(EinFFT), SiMBA(Monarch), and SiMBA(MLP), where Mamba is used for sequence modeling, and EinFFT, Monarch Mixing, and standard MLP are used for channel modeling, respectively. Although there is a slight difference in performance between SiMBA with EinFFT as a channel mixer compared to using MLP as a channel mixer, the MLP version demonstrates superior performance in terms of top-1 accuracy. However, it is noted that the MLP version is less efficient in terms of parameters and FLOPS. The SiMBA (MLP) variant outperforms other SSM models in terms of top-1 accuracy, parameters, and FLOPS.

Moving to the base size, SiMBA-B demonstrates superior performance compared to other SSMs such as Hyena-ViT-B (78.5\%), Vim-S (80.5\%), VMamba-S (83.5\%), and S4ND-ViT-B (80.4\%), while maintaining similar parameters and GFlops. SiMBA-B achieves an impressive accuracy of 84.7\%, which is on par with transformer-based models like Wave-ViT-B (84.8\%), iFormer-B (84.6\%), and Max-ViT-S (84.5\%), but with notably fewer parameters and GFlops. Notably, the SiMBA (MLP) variant outperforms other SSM models and the current state-of-the-art model in terms of top-1 accuracy, while also reducing the number of parameters and FLOPS, thus marking a significant advancement in state-of-the-art performance. 

In the large category, SiMBA-L achieves an 83.9\% top-1 accuracy, surpassing VMamba-B (83.2\%) with comparable computational complexity and parameters. However, a performance gap remains compared to state-of-the-art transformers such as SVT-H-L (85.7\%). Future work aims to address this gap by further scaling SiMBA. However, SiMBA-L (EinFFT) outperforms other SSMs in this range.  Notably, the SiMBA(MLP) variant exhibits instability issues in large network sizes, prompting us to solely report results from the SiMBA(EinFFT) variant for large network sizes. SiMBA(EinFFT) mitigates stability concerns at large network sizes and demonstrates greater efficiency in terms of FLOPS and parameters, outperforming all other SSMs, while with certain trade-offs.

\begin{table}[tp]
\linespread{1.2}
	\centering
	\resizebox{\linewidth}{!}{
		\begin{tabular}{cc|c|cc|cc|cc|cc|cc|cc|cc|cc|cc|ccc}
			\cline{2-23}
			&\multicolumn{2}{c|}{Models}& \multicolumn{2}{c|}{SiMBA}& \multicolumn{2}{c|}{TimesNet}& \multicolumn{2}{c|}{Crossformer}& \multicolumn{2}{c|}{PatchTST}& \multicolumn{2}{c|}{ETSFormer}& \multicolumn{2}{c|}{DLinear}& \multicolumn{2}{c|}{FEDFormer}& \multicolumn{2}{c}{Autoformer}&\multicolumn{2}{c}{Pyraformer}&\multicolumn{2}{c}{MTGNN}& \\
			\cline{2-23}
			&\multicolumn{2}{c|}{Metric}&MSE&MAE&MSE&MAE&MSE&MAE&MSE&MAE&MSE&MAE&MSE&MAE&MSE&MAE&MSE&MAE&MSE&MAE&MSE&MAE\\
			\cline{2-23}
			&\multirow{4}*{\rotatebox{90}{ETTm1}}& 96 &\textbf{0.324}&\textbf{0.360} & \uline{0.338} & \uline{0.375} & 0.349 & 0.395 & 0.339 & 0.377 & 0.375 & 0.398 & 0.345 & 0.372 & 0.379 & 0.419 & 0.505 & 0.475 & 0.543 & 0.510 & 0.379 & 0.446 \\
            &\multicolumn{1}{c|}{}& 192 & \textbf{0.363}&\textbf{0.382} & \uline{0.374} & \uline{0.387} & 0.405 & 0.411 & 0.376 & 0.392 & 0.408 & 0.410 & 0.380 & 0.389 & 0.426 & 0.441 & 0.553 & 0.496 & 0.557 & 0.537 & 0.470 & 0.428 \\
            &\multicolumn{1}{c|}{}& 336 & \textbf{0.395} &\textbf{0.405}  & 0.410 & \uline{0.411} & 0.432 & 0.431 & \uline{0.408} & 0.417 & 0.435 & 0.428 & 0.413 & 0.413 & 0.445 & 0.459 & 0.621 & 0.537 & 0.754 & 0.655 & 0.473 & 0.430 \\
            &\multicolumn{1}{c|}{}& 720 & \textbf{0.451} &\textbf{0.437}  & \uline{0.478} & \uline{0.450} & 0.487 & 0.463 & 0.499 & 0.461 & 0.499 & 0.462 & 0.474 & 0.453 & 0.543 & 0.490 & 0.671 & 0.561 & 0.908 & 0.724 & 0.553 & 0.479 \\
			\cline{2-23}
			&\multirow{4}*{\rotatebox{90}{ETTm2}}& 96 &\textbf{0.177} & \textbf{0.263}& \uline{0.187}& \uline{0.267}& 0.208& 0.292& 0.192& 0.273& 0.189 & 0.280 & 0.193 & 0.292 & 0.203 & 0.287 & 0.255 & 0.339 & 0.435 & 0.507 & 0.203 & 0.299 \\
            &\multicolumn{1}{c|}{} & 192 &\textbf{0.245} &\textbf{0.306} & \uline{0.249}& \uline{0.309}& 0.263& 0.332& 0.252& 0.314& 0.253 & 0.319 & 0.284 & 0.362 & 0.269 & 0.328 & 0.281 & 0.340 & 0.730 & 0.673 & 0.265 & 0.328 \\ 
            &\multicolumn{1}{c|}{}& 336 &\textbf{0.304} &\textbf{0.343} & 0.321& \uline{0.351}& 0.337& 0.369& \uline{0.318}& 0.357& 0.314 & 0.357 & 0.369 & 0.427 & 0.325 & 0.366 & 0.339 & 0.372 & 1.201 & 0.845 & 0.365 & 0.374 \\
            &\multicolumn{1}{c|}{}&720 & \textbf{0.400}&\textbf{0.399} &\uline{ 0.408}& \uline{0.403}& 0.429& 0.430& 0.413& 0.416& 0.414 & 0.413 & 0.554 & 0.522 & 0.421 & 0.415 & 0.433 & 0.432 & 3.625 & 1.451 & 0.461 & 0.459 \\
            \cline{2-23}
			&\multirow{4}*{\rotatebox{90}{ETTh1}}& 96 & \textbf{0.379} & \textbf{0.395} & \uline{0.384} & \uline{0.402} & 0.384 & 0.428 & 0.385 & 0.408 & 0.494 & 0.479 & 0.386 & 0.400 & 0.376 & 0.419 & 0.449 & 0.459 & 0.664 & 0.612 & 0.515 & 0.517 \\
			&\multicolumn{1}{c|}{}&192 & \uline{0.432} &\textbf{0.424 } & 0.436 & \uline{0.429} & 0.438 & 0.452 &\textbf{ 0.431} & 0.432 & 0.538 & 0.504 & 0.437 & 0.432 & 0.420 & 0.448 & 0.500 & 0.482 & 0.790 & 0.681 & 0.553 & 0.522 \\
			&\multicolumn{1}{c|}{}& 336 & \textbf{0.473} & \textbf{0.443} & 0.491 & 0.469 & 0.495 & 0.483 & \uline{0.485} & \uline{0.462} & 0.574 & 0.521 & 0.481 & 0.459 & 0.459 & 0.465 & 0.521 & 0.496 & 0.891 & 0.738 & 0.612 & 0.577 \\
			&\multicolumn{1}{c|}{}& 720 & \textbf{0.483} &\textbf{0.469}  & 0.521 & 0.500 & 0.522 & 0.501 & \uline{0.497} & \uline{0.483} & 0.562 & 0.535 & 0.519 & 0.516 & 0.506 & 0.507 & 0.514 & 0.512 & 0.963 & 0.782 & 0.609 & 0.597 \\
			\cline{2-23}
			&\multirow{4}*{\rotatebox{90}{ETTh2}}& 96 & \textbf{0.290} & \textbf{0.339} & \uline{0.340} & \uline{0.374} & 0.347 & 0.391 & 0.343 & 0.376 & 0.340 & 0.391 & 0.333 & 0.387 & 0.358 & 0.397 & 0.346 & 0.388 & 0.645 & 0.597 & 0.354 & 0.454 \\ 
            &\multicolumn{1}{c|}{} & 192 & \textbf{0.373} &\textbf{ 0.390} & \uline{0.402} & \uline{0.414} & 0.419 & 0.427 & 0.405 & 0.417 & 0.430 & 0.439 & 0.477 & 0.476 & 0.429 & 0.439 & 0.456 & 0.452 & 0.788 & 0.683 & 0.457 & 0.464 \\
            &\multicolumn{1}{c|}{}& 336 & \textbf{0.376} & \textbf{0.406} & 0.452 & \uline{0.452} & 0.449 & 0.465 &\uline{ 0.448} & 0.453 & 0.485 & 0.479 & 0.594 & 0.541 & 0.496 & 0.487 & 0.482 & 0.486 & 0.907 & 0.747 & 0.515 & 0.540 \\
            &\multicolumn{1}{c|}{}& 720 &\textbf{0.407}  & \textbf{0.431} & \uline{0.462} & 0.468 & 0.479 & 0.505 & 0.464 & \uline{0.483} & 0.500 & 0.497 & 0.831 & 0.657 & 0.463 & 0.474 & 0.515 & 0.511 & 0.963 & 0.783 & 0.532 & 0.576 \\
			\cline{2-23}
			&\multirow{4}*{\rotatebox{90}{Electricity}}& 96 & \uline{0.165} &\textbf{ 0.253} & 0.168 & 0.272 & 0.185 & 0.288 & \textbf{0.159} & \uline{0.268} & 0.187 & 0.304 & 0.197 & 0.282 & 0.193 & 0.308 & 0.201 & 0.317 & 0.386 & 0.449 & 0.217 & 0.318\\
            &\multicolumn{1}{c|}{}& 192 &\textbf{0.173 } & \textbf{0.262} & 0.184& 0.289& 0.201& 0.295& \uline{0.177}& \uline{0.278}& 0.199& 0.315& 0.196& 0.285& 0.201& 0.315& 0.222& 0.334& 0.378& 0.443& 0.238& 0.352 \\
            &\multicolumn{1}{c|}{}& 336 & \textbf{0.188} & \textbf{0.277} & 0.198 & 0.300 & 0.211 & 0.312 & \uline{0.195} & \uline{0.296} & 0.212 & 0.329 & 0.209 & 0.301 & 0.214 & 0.329 & 0.231 & 0.338 & 0.376 & 0.443 & 0.260 & 0.348 \\
            &\multicolumn{1}{c|}{}& 720 & \textbf{0.214} & \textbf{0.305} & 0.220 & 0.320 & 0.223 & 0.335 & \uline{0.215} & \uline{0.317} & 0.233 & 0.345 & 0.245 & 0.333 & 0.246 & 0.355 & 0.254 & 0.361 & 0.376 & 0.445 & 0.290 & 0.369 \\
			\cline{2-23}
			&\multirow{4}*{\rotatebox{90}{Traffic}}& 96 & \textbf{0.468} & \textbf{0.268} & 0.593 & 0.321 & 0.591 & 0.329 &  \uline{0.583} &  \uline{0.319} & 0.607 & 0.392 & 0.650 & 0.396 & 0.587 & 0.366 & 0.613 & 0.388 & 0.867 & 0.468 & 0.660 & 0.437 \\
            &\multicolumn{1}{c|}{}&192 &\textbf{0.413}  &\textbf{ 0.317} & 0.617 & 0.336 & 0.607 & 0.345 & \uline{ 0.591} &  \uline{0.331} & 0.621 & 0.399 & 0.598 & 0.370 & 0.604 & 0.373 & 0.616 & 0.382 & 0.869 & 0.467 & 0.649 & 0.438 \\
            &\multicolumn{1}{c|}{}& 336 & \textbf{0.529} & \textbf{0.284} & 0.629 & 0.336 & 0.613 & 0.339 &  \uline{0.599} &  \uline{0.332} & 0.622 & 0.396 & 0.605 & 0.373 & 0.621 & 0.383 & 0.622 & 0.337 & 0.881 & 0.469 & 0.653 & 0.472 \\
            &\multicolumn{1}{c|}{}& 720 & \textbf{0.564} & \textbf{0.297} & 0.640 & 0.350 & 0.620 & 0.348 &  \uline{0.601} &  \uline{0.341} & 0.632 & 0.396 & 0.645 & 0.394 & 0.626 & 0.382 & 0.660 & 0.408 & 0.896 & 0.473 & 0.639 & 0.437 \\
			\cline{2-23}
			&\multirow{4}*{\rotatebox{90}{Weather}}&96 & 0.176 & \textbf{0.219} & \uline{0.172} & \uline{0.220} & 0.191 & 0.251 & \textbf{0.171} & 0.230 & 0.197 & 0.281 & 0.196 & 0.255 & 0.217 & 0.296 & 0.266 & 0.336 & 0.622 & 0.556 & 0.230 & 0.329 \\ 
            &\multicolumn{1}{c|}{}& 192 & 0.222 & \textbf{0.260} & \textbf{0.219} & \uline{0.261} & \uline{0.219} & 0.279 & 0.219 & 0.271 & 0.237 & 0.312 & 0.237 & 0.296 & 0.276 & 0.336 & 0.307 & 0.367 & 0.739 & 0.624 & 0.263 & 0.322 \\
            &\multicolumn{1}{c|}{}& 336 & \textbf{0.275} & \textbf{0.297} &  \uline{0.280} &  \uline{0.306} & 0.287 & 0.332 & 0.277 & 0.321 & 0.298 & 0.353 & 0.283 & 0.335 & 0.339 & 0.380 & 0.359 & 0.395 & 1.004 & 0.753 & 0.354 & 0.396 \\ 
            &\multicolumn{1}{c|}{}&720 &\textbf{ 0.350} & \textbf{0.349} &  \uline{0.365} &  \uline{0.359} & 0.368 & 0.378 & 0.365 & 0.367 & 0.352 & 0.288 & 0.345 & 0.381 & 0.403 & 0.428 & 0.419 & 0.428 & 1.420 & 0.934 & 0.409 & 0.371 \\ 
			\cline{2-23}
		\end{tabular}
	}
 \caption{Multivariate long-term forecasting results with supervised SiMBA. We use prediction lengths  $T \in \{96, 192, 336, 720\}$ for all the datasets for lookup window 96. The best results are in \textbf{bold} and the second best is \uline{underlined}.}
\label{tab:SiMBA_MTS}
 \vspace{-0.4in}
\end{table}

\subsection{SOTA for Multi-Variate Time Series Forecasting}

We conducted a comprehensive evaluation of our State Space model, SiMBA, on seven benchmark standard datasets widely used for Multivariate Time Series Forecasting, including Electricity, Weather, Traffic, and four ETT datasets (ETTh1, ETTh2, ETTm1, and ETTm2), as presented in Table \ref{tab:SiMBA_MTS}. Our evaluation compares SiMBA with various state-of-the-art models, including Transformer-based methods like PatchTST \cite{nie2022time}, CrossFormer \cite{zhang2022crossformer}, FEDFormer \cite{zhou2022fedformer}, ETSFormer \cite{woo2022etsformer}, PyraFormer \cite{liu2021pyraformer}, and AutoFormer \cite{chen2021autoformer}. Additionally, CNN-based methods such as TimeNet \cite{wu2022timesnet}, graph-based methods like MTGNN \cite{wu2020connecting}, and MLP-based models like DLinear \cite{zeng2023transformers} are included in the comparison.

SiMBA demonstrates superior performance across multiple evaluation metrics, including Mean Squared Error (MSE) and Mean Absolute Error (MAE), outperforming the state-of-the-art models. These results underscore the versatility and effectiveness of the SiMBA architecture in handling diverse time series forecasting tasks and modalities, solidifying its position as a leading model in the field. While presenting our results, it's important to note that due to space limitations in Table \ref{tab:SiMBA_MTS}, we couldn't include some recent methods in the Time series domain, such as FourierGNN \cite{yi2023fouriergnn}, CrossGNN \cite{huang2023crossgnn}, TiDE \cite{das2023long}, SciNet \cite{liu2022scinet}, and FreTS \cite{yi2024frequency}. For a fair comparison, we utilized the code from PatchTST \cite{nie2022time}, and the results were based on a lookup window of size 96 for all datasets.


\begin{table}[htb]
    \centering
    \begin{tabular}{lccccc} \toprule
    {Model}& Param  & Top-1\% & Top-5\% & Description(Seq-mix, Channel-mix)\\ \midrule
    ResNet-152& 60M & 78.6 & 94.3 & ConvNet, MLP \\ 
    ViT-b &87M & 78.5 & 93.6 & Attention, MLP \\\midrule
    s4 only$^\dagger$ &13.2M& 58.9 & 82.5 & S4, NA\\
    Hippo only$^\dagger$ &16.4M& 63.2 & 89.2 & Hippo, NA \\
    Mamba only$^\dagger$ &15.3M& 39.1 & 67.1 & Mamba, NA \\
    \midrule
        s4+MLP  &23.5M& 75.9 & 93.6 & S4, MLP\\
    Hippo+MLP  &24.8M& 77.9 & 94.0 & Hippo, MLP \\   \midrule
     SiMBA( Mamba +Monach) &18.5M& 81.1 & 95.5 & Mamba, Monach \\
     SiMBA(Mamba +EinFFT) &15.3M& 81.7 & 95.9 & Mamba, EinFFT \\  
    SiMBA( Mamba +MLP) &26.6M& 84.0 & 96.7 & Mamba, MLP \\
    \bottomrule 
    \end{tabular} 
    \caption{Ablation Analysis on ImageNet-1k for small size model. $^\dagger$ indicates that instability is encountered during the training of the SSMs}    
    \label{tab:ablation_SiMBA}
    \vspace{-0.4in}
\end{table}

\subsection{Ablation Analysis of Model SiMBA}



In our ablation analysis on the ImageNet-1k dataset for small-sized models (Table \ref{tab:ablation_SiMBA}), we explored various sequence modeling configurations, focusing on SiMBA and comparing it with other state space models (SSMs) like S4 and Hyena. Notably, Mamba alone faced stability issues and was non-trainable for the ImageNet dataset for large-scale networks, prompting us to integrate it with different channel mixing components. The results highlight that both S4 and Hyena when utilized as standalone models without channel mixing, exhibit a performance gap compared to state-of-the-art vision transformers like SVT, WaveViT, Volo, and MaxViT.
To address this, our SiMBA architecture combines Mamba as the sequence modeling component with different channel mixing strategies. In the ablation study for channel mixing, we explored three components: MLP, Monarch Mixing (M2), and our  EinFFT. The findings in Table \ref{tab:ablation_SiMBA} indicate that EinFFT-based SiMBA is a channel modeling component which is an alternative method to solve the stability issue, showcasing superior performance with other SSMs when coupled with Mamba as the sequence modeling component. This analysis underscores the significance of channel mixing strategies in enhancing the effectiveness of SiMBA on ImageNet-1k, providing valuable insights for optimizing small-sized models in sequence modeling tasks.

\begin{table}[t]
\centering
\setlength{\tabcolsep}{0.2cm}
\begin{tabular}{c|ccc|ccc|cc}
\toprule
\multicolumn{9}{c}{\textbf{Mask R-CNN 1$\times$ schedule}}\\
\midrule
Backbone & AP$^\text{b}$ & AP$^\text{b}_\text{50}$ & AP$^\text{b}_\text{75}$ & AP$^\text{m}$ & AP$^\text{m}_\text{50}$ & AP$^\text{m}_\text{75}$ & \#param. & FLOPs \\

\midrule

ResNet-101 & 38.2 & 58.8 & 41.4 & 34.7 & 55.7 & 37.2 & 63M & 336G \\
Swin-S & 44.8 & 66.6 & 48.9 & 40.9 & 63.2 & 44.2 & 69M & 354G \\
ConvNeXt-S & 45.4 & 67.9 & 50.0 & 41.8 & 65.2 & 45.1 & 70M & 348G \\
PVTv2-B3 & 47.0 & 68.1 & 51.7 & 42.5 & 65.7 & 45.7 & 65M & 397G \\
\rowcolor{gray!15} SiMBA-S& 46.9 & \textbf{68.6}  & \textbf{51.7}  & \textbf{42.6} & \textbf{65.9}  &\textbf{ 45.8}  & 60M & 382G\\

\bottomrule
\end{tabular}
\caption{\textbf{Object detection and instance segmentation results on COCO dataset}. The performances of various vision models on the COCO val2017 dataset for the downstream tasks of object detection and instance segmentation. RetinaNet is used as the object detector for the object detection task, and the Average Precision ($AP$) at different IoU thresholds or two different object sizes (\emph{i.e.}, small and base) are reported for evaluation. For instance segmentation task, we adopt Mask R-CNN as the base model, and the bounding box and mask Average Precision (\emph{i.e.}, $AP^b$ and $AP^m$) are reported for evaluation. "$1\times$" indicates models fine-tuned for 12 epochs.
}
\label{tab:task_learning_1}
 \vspace{-0.25in}
\end{table}

 \begin{table}[htbp]
 \vspace{-0.1in}
 \scriptsize
  \centering
  \caption{\textbf{Results on transfer learning datasets}. We report the top-1 accuracy on the four datasets as well as the number of parameters and FLOPs. } 
  \vspace{-0.1in}
    \begin{tabular}{l| cccc}
    \toprule
Model  &   {CIFAR-10}   & {CIFAR-100} & {Flowers-102} & {Cars-196} \\ 
    \midrule
    ResNet50~\cite{he2016deep})    & - & - & 96.2  & 90.0 \\
    EfficientNet-B7~\cite{tan2019efficientnet} & 98.9  & 91.7  & 98.8  & 92.7 \\
    ViT-B/16~\cite{dosovitskiy2020image}  & 98.1  & 87.1  & 89.5  & - \\
    ViT-L/16~\cite{dosovitskiy2020image}       & 97.9  & 86.4  & 89.7  & - \\
    Deit-B/16~\cite{touvron2021training}     & \textbf{99.1}  & \textbf{90.8}  & 98.4  & 92.1 \\
    ResMLP-24~\cite{touvron2022resmlp}    & 98.7  & 89.5  & 97.9  & 89.5 \\       
     
     GFNet-XS~(\cite{rao2021global})   &  98.6  &  89.1  &   98.1  &  92.8 \\
     GFNet-H-B~(\cite{rao2021global})   &  99.0  &  90.3  &   98.8  & 93.2 \\\midrule
     SiMBA-B   &  98.7  &  89.3  &   {98.4}  &  {92.7} \\
     \bottomrule
    \end{tabular}%
  \label{tab:transfer_learning}%
\end{table}%

\subsection{Task Learning: Object Detection}

In our experiments on the MS COCO 2017 dataset, a widely used benchmark for object detection and instance segmentation, consisting of approximately 118,000 training images and 5,000 validation images, we employed two widely-used detection frameworks: RetinaNet \cite{lin2017focal} and Mask R-CNN \cite{he2017mask}. Model performance was assessed using Average Precision (AP). SiMBA-s, pre-trained on the ImageNet-1K dataset, served as the backbone architecture, and Xavier initialization was applied to additional layers. The results, detailed in Table \ref{tab:task_learning_1}, showcase SiMBA's competitive performance in comparison to RetinaNet \cite{lin2017focal} and Mask R-CNN \cite{he2017mask}. Notably, SiMBA surpasses the latest models, including LITv2 \cite{panfast}, RegionViT, and PVT \cite{wang2021pyramid} transformer models, outperforming ResNet in terms of AP. SiMBA exhibits superior performance over vanilla ViT models and hierarchical transformer models, achieving the highest AP. 
Furthermore, we extend SiMBA's capabilities to semantic segmentation on the ADE20K dataset using UperNet \cite{xiao2018unified}, with results summarized in Table \ref{tab:exp_ade20k}. The outcomes underscore SiMBA's versatility and effectiveness across diverse computer vision tasks.

\vspace{-0.1in}
 \subsection{Transfer Learning Comparison}

To evaluate the efficacy of the SiMBA architecture and learned representations, we conducted assessments on various transfer learning benchmark datasets, including CIFAR-10~\cite{krizhevsky2009learning}, CIFAR-100~\cite{krizhevsky2009learning}, Stanford Cars~\cite{krause20133d}, and Flowers-102~\cite{nilsback2008automated}. The performance of SiMBA, pre-trained on ImageNet-1K and fine-tuned for image classification on these datasets, was compared against ResMLP models and state-of-the-art models such as GFNet \cite{rao2021global}. Table \ref{tab:transfer_learning} highlights the competitive performance of SiMBA on downstream datasets, outperforming ResMLP models and demonstrating comparable results to GFNet\cite{rao2021global}. Additionally, SiMBA's performance was competitive against popular models like ViT-b, HyenaViT-b, and ViT-b-Monarch on ImageNet-1k. Notably, SiMBA achieved superior results compared to ResNet-152, despite having fewer parameters, emphasizing its effectiveness in transfer learning scenarios.

\begin{table}[t]
\centering
\setlength{\tabcolsep}{0.2cm}
\begin{tabular}{c|c|cc|ccc}
\toprule
Method &  size & mIoU (SS) & mIoU (MS) & \#Param. & FLOPs \\
\midrule
\midrule
ResNet-101 & $512^{2}$ & 42.9 & 44.0 & 85M & 1030G  \\
DeiT-B + MLN & $512^{2}$ & 45.5 & 47.2 & 144M & 2007G \\
Swin-S & $512^{2}$ & 47.6 & 49.5 & 81M & 1039G \\
ConvNeXt-S & $512^{2}$ & 48.7 & 49.6 & 82M & 1027G \\
\rowcolor{gray!15}
SiMBA-S & $512^{2}$ & 49.0 & 49.6 & 62M & 1040G  \\
\bottomrule
\end{tabular}
\caption{\textbf{Semantic segmentation results on ADE20K using UperNet~\cite{xiao2018unified}}.
We evaluate the performance of semantic segmentation on the ADE20K dataset with UperNet~\cite{xiao2018unified}. The FLOPs are calculated with input sizes of $512\times2048$ based on the crop size. "SS" and "MS" denote single-scale and multi-scale testing, respectively.
}
\label{tab:exp_ade20k}
 \vspace{-0.3in}
\end{table}

\section{Conclusion }
This paper has proposed a new channel modeling technique EinFFT which solves the key stability problem in Mamba. We have also proposed SiMBA, a new architecture that leverages EinFFT for channel modeling and Mamba for sequence modeling. SiMBA allows for the exploration of various alternatives for sequence modeling like S4, long conv, Hyena, H3, RWKV, and even newer state space models. Importantly, SiMBA allows the exploration of alternatives for channel modeling like M2, and EinFFT as well as other possible spectral techniques. SiMBA also bridges the performance gap that most state space models have with state-of-art transformers on both vision and time series datasets. We plan to explore a few alternatives within the SiMBA framework such as long conv for sequence modeling with M2 or EinFFT for channel modeling.

\bibliographystyle{splncs04}
\bibliography{main}

\appendix

\section{Appendix }
This section includes comprehensive details regarding training configurations for both transfer learning and task learning tasks. Specifically, Table-\ref{tab:transfer_learning_dataset} provides a breakdown of the dataset information utilized for transformer learning, offering insights into the datasets employed in the training process. For a thorough understanding of the SiMBA algorithm, Algorithm-\ref{alg:block} is presented, providing a complete architectural view of the SiMBA model. This algorithm outlines the key operations and processes involved in the SiMBA architecture, facilitating a deeper understanding of its inner workings. Furthermore, detailed implementation insights into the EinFFT channel mixing method are provided, shedding light on the mechanism utilized for frequency-domain channel mixing within the SiMBA model. This implementation detail offers clarity on how the frequency-domain information is leveraged to enhance the model's performance in capturing intricate relationships within sequential data.
This comparison aids in understanding the different approaches employed by various vision models in handling sequential data and exploiting inter-channel relationships. Moreover, the document reports the object detection performance of two prominent models, GFL~\cite{li2020generalized} and Cascade Mask R-CNN~\cite{cai2018cascade}, on the MS COCO val2017 dataset. The performance metrics presented in Table-~\ref{tab:task_learning_2} demonstrate a notable improvement in performance, showcasing the effectiveness of the proposed SiMBA model.

\begin{figure}[htb]
   
        \centering
        \includegraphics[width=0.47943\textwidth]{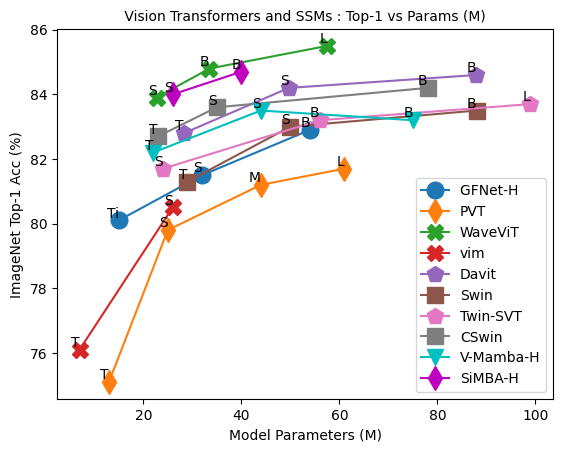}
            \includegraphics[width=0.47943\textwidth]{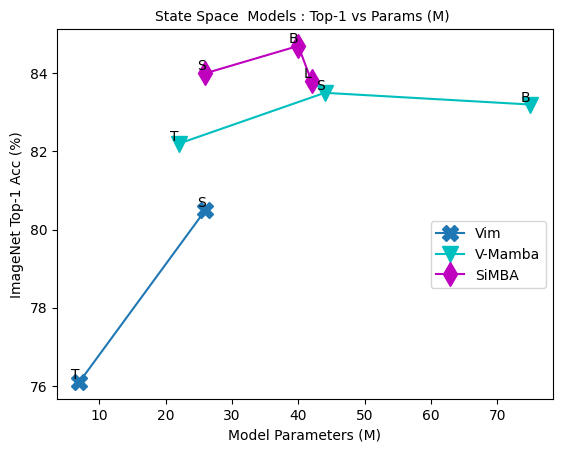}
            \caption{Comparison of ImageNet Top-1 Accuracy (\%) vs Params(M) of Transformer architectures and State space architectures. SiMBA is a better state space model compared to other state space models for Vision data. }
            \label{fig:gflops}
\end{figure}

\begin{figure}[htb]
   
        \centering
        \includegraphics[width=0.47943\textwidth]{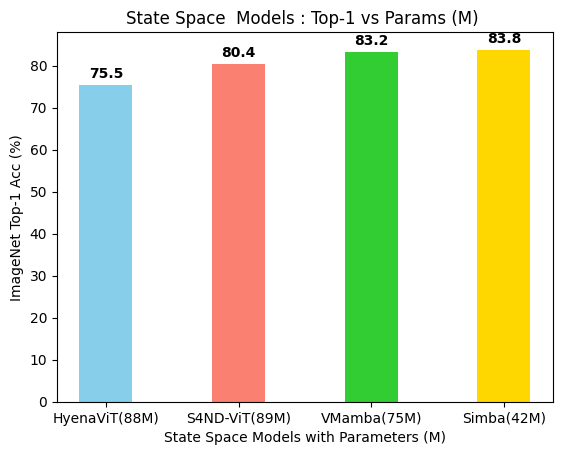}
            \caption{Comparison of ImageNet Top-1 Accuracy (\%) vs Params(M) of State space architectures. SiMBA is a better state space model compared to other state space models for Vision data. }
            \label{fig:params}
\end{figure}


\section{EinFFT Implementation}

We proposed EinFFT module introduces a novel approach to model channels more efficiently, with complex-valued weights and advanced signal processing techniques. The architecture leverages Fast Fourier Transform (FFT) operations, which transform the physical domain to the Frequency Domain. The module's distinctive formulation employs complex-valued weights initialized with a scale factor, promoting enhanced representation learning. The incorporation of soft thresholding introduces sparsity constraints, facilitating noise reduction and feature selection. The complex-valued weights contribute to a more expressive model, enriching the representational space. The modular and extensible implementation of these operations in the code presents a valuable tool for researchers and practitioners in sequence modeling, offering a unique perspective on capturing intricate dependencies within sequential data.

Mathematically, the core computations of EinFFT can be represented as follows. Let \( x \) denote the input tensor of shape \( B \times N \times C \), where \( B \) is the batch size, \( N \) is the sequence length, and \( C \) is the number of channels. The FFT operation is denoted as \( \text{FFT}(x, \text{dim}=1, \text{norm}='ortho') \). Complex-valued weights \( W_1, W_2, B_1, B_2 \) are utilized in a series of operations involving complex multiplication, rectified linear units (ReLU), and soft thresholding:

\begin{align*}
X &= \text{FFT}(x) \\
X_{\text{real}_1} &= \max(\text{Re}(X) \cdot W_{1, \text{real}} - \text{Im}(X) \cdot W_{1, \text{imag}} + B_{1, \text{real}}, 0) \\
X_{\text{imag}_1} &= \max(\text{Re}(X) \cdot W_{1, \text{imag}} + \text{Im}(X) \cdot W_{1, \text{real}} + B_{1, \text{imag}}, 0) \\
X_{\text{real}_2} &= \text{Re}(X_{\text{real}_1} \cdot W_{2, \text{real}} - X_{\text{imag}_1} \cdot W_{2, \text{imag}} + B_{2, \text{real}}) \\
X_{\text{imag}_2} &= \text{Im}(X_{\text{real}_1} \cdot W_{2, \text{imag}} + X_{\text{imag}_1} \cdot W_{2, \text{real}} + B_{2, \text{imag}}) \\
X_{\text{shrink}} &= \text{softshrink}(X_{\text{real}_2} \, \& \, X_{\text{imag}_2}, \lambda=\text{sparsity\_threshold}) \\
x_{\text{ifft}} &= \text{IFFT}(X_{\text{shrink}}, \text{dim}=1, \text{norm}='ortho') \\
x_{\text{reshaped}} &= \text{reshape}(x_{\text{ifft}}, (B, N, C))
\end{align*}

This sequence of operations encapsulates the essence of the EinFFT module, providing a comprehensive mathematical description of its functionality.

\begin{algorithm}[h]
\caption{SiMBA Block Process}
\small
\begin{algorithmic}[1]
\STATE{\textbf{Input:} Patch sequence $\mathbf{X}_{l-1}$ : \textcolor{shapecolor}{$(\mathtt{B}, \mathtt{N}, \mathtt{D})$}}
\STATE{\textbf{Output:} Patch sequence $\mathbf{X}_{l}$ : \textcolor{shapecolor}{$(\mathtt{B}, \mathtt{N}, \mathtt{D})$}}
\STATE \textcolor{gray}{\text{/* normalize the input sequence $\mathbf{X}_{l-1}'$ */}}
\STATE $\mathbf{X}_{l-1}'$ : \textcolor{shapecolor}{$(\mathtt{B}, \mathtt{N}, \mathtt{D})$} $\leftarrow$ $\mathbf{Norm}(_{l-1})$
\STATE $\mathbf{x}$ : \textcolor{shapecolor}{$(\mathtt{B}, \mathtt{N}, \mathtt{P})$} $\leftarrow$ $\mathbf{Linear}^\mathbf{x}(\mathbf{X}_{l-1}')$
\STATE $\mathbf{z}$ : \textcolor{shapecolor}{$(\mathtt{B}, \mathtt{N}, \mathtt{P})$} $\leftarrow$ $\mathbf{Linear}^\mathbf{z}(\mathbf{X}_{l-1}')$
\STATE \textcolor{gray}{\text{/* process with different direction */}}
\FOR{$o$ in \{loop\}}
    \STATE $\mathbf{x}'_o$ : \textcolor{shapecolor}{$(\mathtt{B}, \mathtt{N}, \mathtt{P})$} $\leftarrow$ $\mathbf{SiLU}(\mathbf{Conv1d}_{o}(\mathbf{x}))$
    \STATE $\mathbf{B}_o$ : \textcolor{shapecolor}{$(\mathtt{B}, \mathtt{N}, \mathtt{K})$} $\leftarrow$ $\mathbf{Linear}^{\mathbf{B}}_o(\mathbf{x}'_o)$
    \STATE $\mathbf{C}_o$ : \textcolor{shapecolor}{$(\mathtt{B}, \mathtt{N}, \mathtt{K})$} $\leftarrow$ $\mathbf{Linear}^{\mathbf{C}}_o(\mathbf{x}'_o)$
    \STATE \textcolor{gray}{\text{/* softplus ensures positive $\mathbf{\Delta}_o$ */}}
    \STATE $\mathbf{\Delta}_o$ : \textcolor{shapecolor}{$(\mathtt{B}, \mathtt{N}, \mathtt{P})$} $\leftarrow$ $\log(1 + \exp(\mathbf{Linear}^{\mathbf{\Delta}}_o(\mathbf{x}'_o) + \mathbf{Parameter}^{\mathbf{\Delta}}_o))$
    \STATE \textcolor{gray}{\text{/* shape of $\mathbf{Parameter}^{\mathbf{A}}_o$ is \textcolor{shapecolor}{$(\mathtt{P}, \mathtt{K})$} */}}
    \STATE $\overline{\mathbf{A}}_o$ : \textcolor{shapecolor}{$(\mathtt{B}, \mathtt{N}, \mathtt{P}, \mathtt{K})$} $\leftarrow$ $\mathbf{\Delta}_o \bigotimes \mathbf{Parameter}^{\mathbf{A}}_o$ 
    \STATE $\overline{\mathbf{B}}_o$ : \textcolor{shapecolor}{$(\mathtt{B}, \mathtt{N}, \mathtt{P}, \mathtt{K})$} $\leftarrow$ $\mathbf{\Delta}_o \bigotimes \mathbf{B}_o$
    \STATE $\mathbf{y}_o$ : \textcolor{shapecolor}{$(\mathtt{B}, \mathtt{N}, \mathtt{P})$} $\leftarrow$ $\mathbf{SSM}(\overline{\mathbf{A}}_o, \overline{\mathbf{B}}_o, \mathbf{C}_o)(\mathbf{x}_o')$
\ENDFOR
\STATE \textcolor{gray}{\text{/* get gated $\mathbf{y}_o$ */}}
\STATE $\mathbf{y}_{o}$ : \textcolor{shapecolor}{$(\mathtt{B}, \mathtt{N}, \mathtt{P})$} $\leftarrow$ $\mathbf{y}_{o} \bigodot \mathbf{SiLU}(\mathbf{z}) $
\STATE \textcolor{gray}{\text{/* residual connection */}}
\STATE $\mathbf{X}_{l}$ : \textcolor{shapecolor}{$(\mathtt{B}, \mathtt{N}, \mathtt{D})$} $\leftarrow$ $\mathbf{DP} (\mathbf{Linear}^\mathbf{X}(\mathbf{y}_{o})) + \mathbf{X}_{l-1}$

\STATE $\mathbf{Res}$ : \textcolor{shapecolor}{$(\mathtt{B}, \mathtt{N}, \mathtt{D})$} $\leftarrow$ $ \mathbf{X}_{l}$

\STATE $\mathbf{X}_{l}$ : \textcolor{shapecolor}{$(\mathtt{B}, \mathtt{N}, \mathtt{D})$} $\leftarrow$ $\mathbf{FFT} (\mathbf{X}_{l})$

\STATE $\mathbf{X}_{l}$ : \textcolor{shapecolor}{$(\mathtt{B}, \mathtt{N}, \mathtt{C_b}, \mathtt{C_d})$} $\leftarrow$ $\mathbf{rearrange} (\mathbf{X}_{l}) , \mathtt{D}=\mathtt{C_b} \times \mathtt{C_d}$

\STATE $\mathbf{W}_{1},\mathbf{W}_{2}$ : \textcolor{shapecolor}{$( \mathtt{C_b}, \mathtt{C_d}, \mathtt{C_d})$} $\leftarrow$ {\text{shape of complex value $\mathbf{Parameter}^{\mathbf{W}_1,\mathbf{W}_2}_l$ }}

\STATE $\mathbf{B}_{1},\mathbf{B}_{2}$ : \textcolor{shapecolor}{$( \mathtt{C_b}, \mathtt{C_d})$} $\leftarrow$ {\text{shape of complex value $\mathbf{Parameter}^{\mathbf{B}_1,\mathbf{B}_2}_l$ }}

\STATE $\mathbf{X}_{l}^{Real}$ : \textcolor{shapecolor}{$(\mathtt{B}, \mathtt{N}, \mathtt{C_b}, \mathtt{C_d})$} $\leftarrow$ $ReLU(\mathbf{EMM}(\mathbf{W}_{1},\mathbf{B}_{1}^{Real})(\mathbf{x}_l^{Real}))$

\STATE $\mathbf{X}_{l}^{Img}$ : \textcolor{shapecolor}{$(\mathtt{B}, \mathtt{N}, \mathtt{C_b}, \mathtt{C_d})$} $\leftarrow$ $ReLU(\mathbf{EMM}(\mathbf{W}_{1},\mathbf{B}_{1}^{Img})(\mathbf{x}_l^{Img}))$

\STATE $\mathbf{X}_{l}$ : \textcolor{shapecolor}{$(\mathtt{B}, \mathtt{N}, \mathtt{C_b}, \mathtt{C_d})$} $\leftarrow$ $\mathbf{IFFT}(\mathbf{X}_l^{Real}+j\mathbf{X}_l^{Img})$
\STATE $\mathbf{X}_{l}$ : \textcolor{shapecolor}{$(\mathtt{B}, \mathtt{N}, \mathtt{D})$} $\leftarrow$ $ \mathbf{X}_{l}+ \mathbf{Res}$
\STATE Return: $\mathbf{X}_{l} $ 
\label{alg:block}
\end{algorithmic}
\end{algorithm}

The SiMBA Block Process, depicted in Algorithm~\ref{alg:block} inspire from ViM\cite{zhu2024vision}, unfolds as follows:
Initially, the input token sequence $\mathbf{X}_{\mathtt{l}-1}$ undergoes normalization via a dedicated layer. Subsequently, a linear transformation is employed to project the normalized sequence onto $\mathbf{x}$ and $\mathbf{z}$, both of dimension size $P$. Moving forward, the algorithm proceeds to process $\mathbf{x}$ in both forward and backward directions. Within each direction, a 1-D convolution operation is applied to $\mathbf{x}$ to yield $\mathbf{x}_{o}$. Subsequently, linear projections are applied to $\mathbf{x}_{o}$, resulting in $\mathbf{B}_{o}$, $\mathbf{C}_{o}$, and $\mathbf{\Delta}_{o}$. Utilizing $\mathbf{\Delta}_{o}$, transformations are performed on $\overline{\mathbf{A}}_{o}$ and $\overline{\mathbf{B}}_{o}$. Following this, the SSM mechanism computes $\mathbf{y}_{forward}$ and $\mathbf{y}_{backward}$. Gating by $\mathbf{z}$ is applied to $\mathbf{y}_{o}$, culminating in the generation of the output token sequence $\mathbf{X}_{\mathtt{l}}$. This output sequence is the aggregation of the gated $\mathbf{y}_{o}$ from both directions. In essence, the SiMBA Block Process encompasses a series of operations, including normalization, linear projections, convolution, and attention mechanisms, ultimately leading to the generation of the final token sequence $\mathbf{X}_{\mathtt{l}}$.

\section{Experiment}

\begin{table}[t]
    \caption{SiMBA training settings.}
    \label{tab:vit-training-details}
    \centering
    \begin{tabular}{rcc}
    \toprule
    {} & ImageNet-1k & CIFAR-10 \\
    \midrule
    Optimizer & \multicolumn{2}{c}{AdamW} \\
    Optimizer momentum & \multicolumn{2}{c}{$\beta_1, \beta_2=0.9, 0.999$} \\
    Learning rate schedule & \multicolumn{2}{c}{Cosine decay w/ linear warmup} \\
    Dropout rate & \multicolumn{2}{c}{0.2} \\
    Label smoothing & \multicolumn{2}{c}{0.1} \\
    \midrule
    Image size & 224 x 224 & 32 x 32 \\
    Base learning rate & 1e-3 & \{1e-4, 3e-4, 1e-3\} \\
    Batch size & 128 & 64 \\
    Training epochs & 300 & up to 1000 \\
    Warmup epochs & 10 & 5 \\
    Stochastic depth rate  & 0.1 & \{0, 0.1\} \\
    Weight decay & 0.05 & \{0, 0.1\} \\
    \bottomrule 
    \end{tabular}
\end{table}

\begin{table*}[!tb]
\centering
\caption{The performances of various vision backbones on COCO val2017 dataset for the downstream task of object detection. Four kinds of object detectors, i.e, GFL ~\cite{li2020generalized}, and  Cascade Mask R-CNN~\cite{cai2018cascade} in mmdetection \cite{chen2019mmdetection}, are adopted for evaluation. We report the bounding box Average Precision ($AP^b$) in different IoU thresholds.}
\begin{tabular}{c|c|ccc}
\Xhline{2\arrayrulewidth}
Backbone   & Method & $AP^b$ & $AP^b_{50}$ & $AP^b_{75}$  \\ \hline

ResNet50 \cite{he2016deep}  & \multirow{4}{*}{GFL \cite{li2020generalized}}
   & 44.5 & 63.0  & 48.3  \\
   
Swin-T   \cite{liu2021swin}     &    & 47.6 & 66.8  & 51.7 \\
PVTv2-B2  \cite{wang2022pvt}  &    & 50.2 & 69.4  & 54.7 \\
SiMBA-S (Ours)                &  & \textbf{50.3} & \textbf{69.0} & \textbf{55.1}\\ 

\hline\hline

ResNet50 \cite{he2016deep}  & \multirow{4}{*}{\begin{tabular}[c]{@{}c@{}}Cascade\\ Mask\\ R-CNN\end{tabular} \cite{cai2018cascade}} & 46.3 & 64.3  & 50.5  \\
Swin-T \cite{liu2021swin}    &       & 50.5 & 69.3  & 54.9  \\
PVTv2-B2 \cite{wang2022pvt}  &     & 51.1 & 69.8  & 55.3   \\

SiMBA-S (Ours) &                   &\textbf{ 51.4} & \textbf{70.1}  & \textbf{56.0 }  \\ \Xhline{2\arrayrulewidth}
\end{tabular}
\label{tab:task_learning_2}
\end{table*}

\subsection{ Dataset and Training setup for Image classification task}
We describe the training process of the image recognition task  using the ImageNet1K benchmark dataset, which includes 1.28 million training images and 50K validation images belonging to 1,000 categories. The vision backbones are trained from scratch using data augmentation techniques like RandAug, CutOut, and Token Labeling objectives with MixToken. The performance of the trained backbones is evaluated using both top-1 and top-5 accuracies on the validation set. The optimization process involves using the AdamW optimizer with a momentum of 0.9, 10 epochs of linear warm-up, and 310 epochs of cosine decay learning rate scheduler. The batch size is set to 128 and is distributed on 8 A100 GPUs. The learning rate and weight decay are fixed at 0.00001 and 0.05, respectively.

\subsection{Training setup for Transfer Learning}
To test the effectiveness of our architecture and learned representation, we evaluated vanilla SiMBA on commonly used transfer learning benchmark datasets, including CIFAR-10~\cite{krizhevsky2009learning}, CIFAR100~\cite{krizhevsky2009learning}, Oxford-IIIT-Flower~\cite{nilsback2008automated} and Standford Cars~\cite{krause20133d}. Our approach followed the methodology of previous studies~\cite{tan2019efficientnet,dosovitskiy2020image,touvron2021training,touvron2022resmlp,rao2021global}, where we initialized the model with ImageNet pre-trained weights and fine-tuned it on the new datasets. In Table-\ref{tab:transfer_learning} of the main paper, we have presented a comparison of the transfer learning performance of our basic and best models with state-of-the-art CNNs and vision transformers. The transfer learning setup employs a batch size of 64, a learning rate (lr) of 0.0001, a weight-decay of 1e-4, a clip-grad of 1, and warmup epochs of 5. We have utilized a pre-trained model trained on the Imagenet-1K dataset, which we have fine-tuned on the transfer learning dataset specified in table-\ref{tab:transfer_learning_dataset} for 1000 epochs.

 \subsection{Task Learning: Object Detection}

\textbf{Training setup: }
In this section, we examine the pre-trained SiMBA-H-small behavior on COCO dataset for two downstream tasks that localize objects ranging from bounding-box level to pixel level, \emph{i.e.}, object detection and instance segmentation. Two mainstream detectors,  \emph{i.e.}, RetinaNet \cite{lin2017focal} and Mask R-CNN\cite{he2017mask} as shown in table-8 of the main paper, and two state-of-the-art detectors \emph{i.e.},  GFL ~\cite{li2020generalized}, and  Cascade Mask R-CNN~\cite{cai2018cascade} in mmdetection \cite{chen2019mmdetection} in this supplementary doc.
We are employed for each downstream task, and we replace the CNN backbones in each detector with our SiMBA-H-small for evaluation. Specifically, each vision backbone is first pre-trained over ImageNet1K, and the newly added layers are initialized with Xavier \cite{glorot2010understanding}. Next, we follow the standard setups in \cite{liu2021swin} to train all models on the COCO train2017 ($\sim$118K images). Here the batch size is set as 16, and AdamW \cite{loshchilovdecoupled} is utilized for optimization (weight decay: 0.05, initial learning rate: 0.0001, betas=(0.9, 0.999)).  We used learning rate (lr) configuration with step lr policy, linear warmup at every 500 iterations with warmup ration 0.001. All models are finally evaluated on the COCO val2017 (5K images).  For state-of-the-art models like GFL ~\cite{li2020generalized}, and  Cascade Mask R-CNN~\cite{cai2018cascade}, we utilize 3 $\times$ schedule (\emph{i.e.}, 36 epochs) with the multi-scale strategy for training, whereas for RetinaNet \cite{lin2017focal} and Mask R-CNN\cite{he2017mask} we utilize 1 $\times$ schedule (\emph{i.e.}, 12 epochs).

\begin{table*}[htbp]
  \centering
  \caption{This table presents information about datasets used for transfer learning. It includes the size of the training and test sets, as well as the number of categories included in each dataset such as CIFAR-10~\cite{krizhevsky2009learning}, CIFAR-100~\cite{krizhevsky2009learning},  Flowers-102~\cite{nilsback2008automated}, Stanford Cars~\cite{krause20133d}.  }
\setlength{\tabcolsep}{4.0pt}
    \begin{tabular}{c| c|c|c|c}
    \toprule
    Dataset  &   CIFAR-10 & CIFAR-100& Flowers-102 & Stanford Cars \\\midrule
    Train Size  & 50,000  & 50,000 & 8,144 & 2,040\\
    Test Size  & 10,000  & 10,000 & 8,041  & 6,149\\
    \#Categories & 10& 100 & 196 & 102\\
     \bottomrule
    \end{tabular}%
  \label{tab:transfer_learning_dataset}%
\end{table*}%

\subsection{SiMBA Architecture Details}

The architectural details of the simplified SiMBA-based context can be expressed as follows:

\begin{equation}
\begin{aligned}
\mathbf{X}_l &= \mathbf{DP}(\mathbf{Mamba}(\mathbf{X}_{\mathtt{l}-1})) + \mathbf{X}_{\mathtt{l}-1}, \\
\mathbf{X}_l &= \mathbf{DP}( \mathbf{EinFFT}(\mathbf{ \mathbf{Norm}(\mathbf{X}_{\mathtt{l}})}))+\mathbf{X}_l, \\
\end{aligned}
\end{equation}

Here, $\mathbf{SiMBA}$ represents the proposed Mamba block, $\mathtt{L}$ denotes the number of layers, and $\mathbf{Norm}$ stands for the normalization layer. The process involves applying the Mamba block to the previous time step $\mathbf{X}_{\mathtt{l}-1}$, followed by a residual connection and dropout ($\mathbf{DP}$). The resulting tensor is then normalized ($\mathbf{Norm}$)  and by applying the EinFFT operation subsequently, which is the proposed frequency-domain channel mixing operation. Finally, the tensor undergoes another dropout and is added to the previous state $\mathbf{X}_l$. This overall structure is iteratively applied for multiple layers ($\mathtt{l}$).

In our proposed SiMBA model, we employ a series of operations to process sequential data effectively. Here's a breakdown of the key components and their interactions:
\begin{itemize}

\item Mamba Block ($\mathbf{Mamba}$): The Mamba block is the fundamental building block of our model, responsible for capturing temporal dependencies within the input time series. By applying this block to the previous time step $\mathbf{X}_{\mathtt{l}-1}$, we aim to extract relevant features and patterns crucial for forecasting. The residual connection with dropout ($\mathbf{DP}$) ensures the retention of essential information from preceding steps, enhancing the model's ability to learn complex temporal dynamics.

\item Normalization ($\mathbf{Norm}$): Normalizing the tensor $\mathbf{X}_{\mathtt{L}}^0$ enhances the stability and convergence of the model during training. This step helps mitigate issues related to gradient vanishing or explosion, promoting smoother learning dynamics.

\item Frequency-Domain Channel Mixing ($\mathbf{EinFFT}$): The operation $\mathbf{EinFFT}$ represents a novel approach to exploit frequency-domain information for capturing intricate relationships among different channels within the data. By transforming the normalized tensor into the frequency domain, our model can effectively capture periodic patterns and complex inter-channel dependencies, thereby improving its forecasting capabilities.

\item Dropout ($\mathbf{DP}$): Dropout is applied after the frequency-domain channel mixing to prevent overfitting and enhance the model's generalization ability. This regularization technique aids in robust feature learning by reducing the model's sensitivity to noise in the data.

\item Residual Connection and Iteration: The residual connection adds the result of the dropout operation to the previous state $\mathbf{X}_l$, facilitating iterative refinement of temporal representations across multiple layers ($\mathtt{l}$). This iterative process enables the model to progressively learn hierarchical features and capture increasingly complex temporal dynamics.
 
\end{itemize}
As for the hyperparameters governing our architecture, we define:

- $\mathtt{L}$: The number of Mamba blocks, determining the depth of the model.
- $\mathtt{D}$: The hidden state dimension, representing the dimensionality of the model's hidden states.
- $\mathtt{P}$: The expanded state dimension, determining the dimensionality of the expanded states within the model.
- $\mathtt{N}$: The sequence length of the input data sequence.

our SiMBA architecture leverages Mamba blocks, frequency-domain channel mixing, normalization, and dropout to effectively process sequential data and capture intricate temporal dependencies. The defined hyperparameters control the model's depth, hidden state dimension, expanded state dimension, and attention mechanism dimensionality, thereby shaping its overall characteristics and performance.

\end{document}